\documentclass[conference]{IEEEtran}
\IEEEoverridecommandlockouts
\usepackage{cite}
\usepackage{amsmath,amssymb,amsfonts}
\usepackage{url}
\usepackage{algorithmic}
\usepackage{graphicx}
\usepackage{textcomp}
\usepackage{xcolor}
\def\BibTeX{{\rm B\kern-.05em{\sc i\kern-.025em b}\kern-.08em
    T\kern-.1667em\lower.7ex\hbox{E}\kern-.125emX}}
\begin{document}

\title{Building a Multivariate Time Series Benchmarking Datasets Inspired by Natural Language Processing (NLP)}

\author{\IEEEauthorblockN{1\textsuperscript{st} Mohammad Asif Ibna Mustafa}
\IEEEauthorblockA{\textit{Department of Computation, Information and Technology} \\
\textit{Technical University of Munich}\\
Arcistrasse 21, 81377, Munich \\
asif.mustafa@tum.de}
\and
\IEEEauthorblockN{2\textsuperscript{nd} Ferdinand Heinrich}
\IEEEauthorblockA{\textit{Fraunhofer Institute for Electronic Microsystems and} \\
\textit{Solid State Technologies EMFT}\\
\textit{Machine Learning Enhanced Sensor Systems}\\
Hansastraße 27 d, 80686 München, Germany \\
ferdinand.heinrich@emft.fraunhofer.de}
}

\maketitle

\begin{abstract}
Time series analysis has become increasingly important in various domains, and developing effective models relies heavily on high-quality benchmark datasets. Inspired by the success of Natural Language Processing (NLP) benchmark datasets in advancing pre-trained models, we propose a new approach to create a comprehensive benchmark dataset for time series analysis. This paper explores the methodologies used in NLP benchmark dataset creation and adapts them to the unique challenges of time series data. We discuss the process of curating diverse, representative, and challenging time series datasets, highlighting the importance of domain relevance and data complexity. Additionally, we investigate multi-task learning strategies that leverage the benchmark dataset to enhance the performance of time series models. This research contributes to the broader goal of advancing the state-of-the-art in time series modeling by adopting successful strategies from the NLP domain.

\end{abstract}

\begin{IEEEkeywords}
Time Series, Natural Language Processing, Multi-Task Learning, Benchmarking Time Series Dataset
\end{IEEEkeywords}

\section{Introduction}

The analysis of time series data is important in various fields such as industry and engineering\cite{b1}, medicine and healthcare\cite{b2} and economics and finance\cite{b3, b4}. One such model in time series is TimeGPT which applies machine learning techniques in time series analysis, specifically in forecasting and anomaly detection\cite{b20}. TimeGPT has shown exceptional potential in understanding and predicting temporal data.

However, with the introduction of advanced systems like TimeGPT, there is a need for robust benchmarking frameworks. Existing benchmarks may not fully capture the complexities and abilities of TimeGPT. Therefore, new standards and datasets are needed to evaluate not only the forecasting or anomaly detection accuracy of TimeGPT but also its computational efficiency, scalability, and adaptability across various contexts.

This paper proposes a detailed framework for benchmarking in the time series domain, drawing inspiration from the evolution of benchmarking in natural language processing (NLP). The paper explores the historical development of NLP benchmarking datasets, noting their crucial role in enhancing model performance. It then addresses the distinct challenges of time series data and proposes integrating NLP-inspired methodologies to create an expansive and robust benchmarking dataset. 

Moreover, we also incorporate multitask learning, which involves developing solutions that can handle multiple tasks simultaneously in time series analysis to enhance performance. This technique aims to not only create robust, accurate, and scalable solutions but also drive progress in a diverse range of applications.

In summary, the paper highlights the critical role of developing a comprehensive benchmarking dataset for time series analysis. The development is not just about enhancing forecasting abilities but also about refining anomaly detection and classification capabilities, setting a new standard in the evaluation and advancement of time series models like TimeGPT.

\section{Related Work}


\begin{figure*}[htbp] 
  \centering
  \includegraphics[width=\textwidth, height=1.7in]{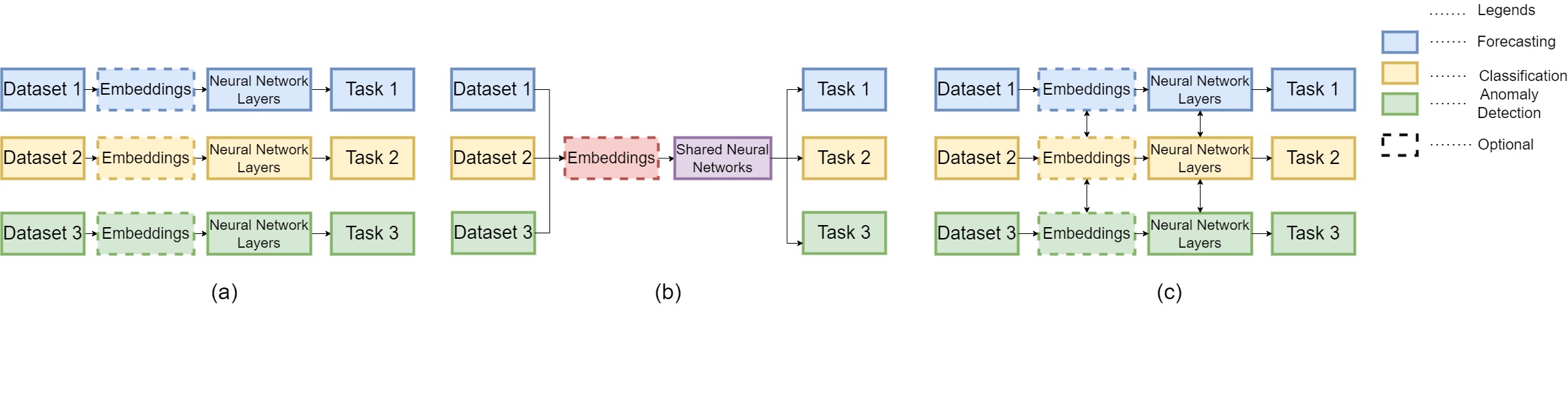} 
  \caption{(a) Single Task Learning, (b) Multi-Task Learning with Hard Parameter Sharing and (c) Multi-Task Learning with Soft Parameter Sharing}
  \label{figure}
\end{figure*}

\subsection{Natural Language Processing}

The NLP community has developed benchmarking initiatives to evaluate language models' performance on various linguistic tasks. One such benchmark is the General Language Understanding Evaluation (GLUE), which focuses on zero-shot inference to assess models' adaptability to unseen data without task-specific training \cite{b10}. GLUE includes tasks such as sentiment analysis, textual entailment, and semantic similarity, providing a standardized framework for a comprehensive evaluation of language models. GLUE's tasks cover a wide range of linguistic challenges and foster a nuanced assessment of models' capabilities in classification, relationship understanding, and more. GLUE selects diverse tasks, collects or creates relevant datasets for each task, and provides clear task definitions and guidelines to ensure consistent annotation. Specific evaluation metrics tailored to each task's nature are used to report the overall performance.

SuperGLUE builds upon GLUE by introducing more challenging NLP tasks that demand advanced reasoning and inference capabilities \cite{b11}. The benchmark employs task-specific metrics, including accuracy, precision, recall, and F1 score, to capture the nuances of each task. The SuperGLUE Benchmark includes tasks that test system understanding and reasoning in English, beyond the current state-of-the-art systems' scope. The benchmark includes a diagnostic dataset that tests models for linguistic, commonsense, and world knowledge. Aggregate system performance is measured by averaging scores across all tasks, with each task weighed equally. Additionally, popular benchmarks like the Stanford Question Answering Dataset (SQuAD) 1.1 and SQuAD 2.0 have played a crucial role in evaluating models' ability to comprehend and answer questions based on given passages \cite{b12}. SQuAD benchmarks, focusing on reading comprehension, have become widely adopted for assessing language models' performance in extracting information and reasoning from textual data.

\subsection{Time Series Analysis}

The most interesting field in time series analysis includes time series classification, time series forecasting, and time series anomaly detection. Time series forecasting has many competitions and datasets to evaluate prediction models such as M3 \cite{b8} and M4 \cite{b9}. These competitions provide large datasets to assess algorithmic performance. Additionally, datasets such as ETTm2, Exchange-Rate, ECL, TrafficL, Weather, and ILI cover various domains such as electricity consumption, financial exchange rates, and healthcare, each posing unique challenges for accurate forecasting. These are the most commonly used datasets among researchers to test time-series benchmarking models \cite{b19, b20, b21, b22, b23}.

There has been a recent review of eight papers introducing innovative methods for multi-class time-series classification in the years 2020-2021 \cite{b13}. In these studies, benchmarking was consistently performed on the 85 datasets from the 2015 version of the UCR archive, primarily comprising fixed-length univariate time series. Among the scrutinized papers, four \cite{b14, b15, b16, b17} exclusively utilized the 2015 version of the UCR archive, with one \cite{b16} introducing an additional synthetic univariate TSC dataset. However, only one study \cite{b15} extended their evaluation to include the 43 datasets newly added in the UCR archive 2018 version. It is worth noting that the benchmark analyses conducted on the archive version of UCR 2018 exclude 16 datasets out of 128 due to their unequal length or missing values.

Time series anomaly detection has gained significant traction in the last five years, becoming a focal point in various data science, database, and machine learning conferences, including SIGKDD \cite{b24, b25}, ICDM \cite{b26}, and VLDB, etc. Researchers in this domain commonly evaluate their methods on popular benchmark datasets, such as those provided by Yahoo \cite{b27}, Numenta \cite{b28}, NASA \cite{b24}, and Pei’s Lab (OMNI) \cite{b25}. This recognition of the intrinsic quality of benchmark datasets contributes to a clearer understanding of progress within the field, fostering more accurate assessments of the efficacy of anomaly detection methods in recent years.

\subsection{Multi-task learning}
Multi-task learning is a technique that trains a model on multiple related tasks simultaneously to improve the model's performance and generalization on each task, as shown in Fig. 1 \cite{b29}. It is often used in natural language processing (NLP) to leverage the shared knowledge and linguistic skills across different NLP tasks, such as sentence classification, natural language inference, question answering, and more. Multi-task learning is used in GLUE \cite{b10} and SuperGLUE \cite{b11}. Multi-task learning can be useful not only for natural language understanding (NLU), but also for time series domain. In time series forecasting, multi-task learning can help with this task because different time series may share common patterns, trends, or seasonality. Additionally, multi-task learning can help to handle missing values, anomalies, or noise in the data. There are various models available that use multi-task learning for time series forecasting\cite{b36, b37}. 


\section{Methodology}

Our research is focused on two datasets that are ideal for time series forecasting and can be used for benchmarking purposes. These datasets are the M4 competition dataset \cite{b9} and the Electricity Consuming Load (ECL) dataset \cite{b32}. The M4 competition dataset \cite{b9} is widely recognized as a benchmark in the field of time-series forecasting and provides a more comprehensive evaluation platform than the M3 competition \cite{b8}. It includes a collection of 100,000 time series from a range of domains, such as finance, economics, demography, and industry. The time series have different periodicities, such as yearly, quarterly, monthly, weekly, daily, and hourly, which helps to identify trends, cycles, and seasonality, and to improve forecasting accuracy. Our experiments use the M4 dataset to ensure a robust evaluation framework that considers general time-series forecasting scenarios and the specific challenges posed by electricity consumption patterns.

The UEA and UCR Time Series Classification Dataset is a well-known benchmarking resource that offers a diverse range of 128 univariate and 30 multivariate time series datasets across various domains\cite{b33}. Its standardized evaluation metric (accuracy) and well-defined train/test split make it an ideal resource for fair and reproducible comparisons among algorithms. Regular updates with new datasets and results from state-of-the-art methods ensure its relevance and accessibility in the research community. We propose leveraging this dataset as our primary benchmarking resource for evaluating proposed time-series classification methods.

We have chosen Yahoo as the benchmarking dataset for time series anomaly detection \cite{b27}. This dataset contains 16 million labelled time series that reflect real-world web traffic from Yahoo \cite{b34}. It is a part of the Yahoo Webscope program, which grants academic researchers access to large-scale datasets. The Yahoo dataset is very popular among researchers and practitioners due to its diversity and size. It covers various types of time series, such as seasonal, trend, and noise. Additionally, it is a realistic and challenging dataset that contains different types of anomalies, including point, contextual, and collective anomalies. What makes this dataset even more valuable is that it provides ground truth labels for the anomalies and their types, which are confirmed by domain experts.


Our benchmarking process focuses on creating a versatile dataset and an effective evaluation framework, leaving model development to the researchers. When it comes to single-task learning, each model is trained exclusively on a specific task, with the goal of optimizing its performance depicted in Fig. 1. In contrast, multi-task learning with hard parameter sharing uses a shared architecture for lower layers across multiple tasks, while having separate, task-specific top layers for each task. On the other hand, multi-task learning with soft parameter sharing involves separate models for each task, but these models are regularized to promote similarity in their parameters, offering a more flexible approach that still benefits from learning across tasks. Another direction for researchers to explore in time series is context-aware learning, which involves the use of embedding. It is up to the researchers to determine whether to use pretrained context-aware learning for model development in time series like GloVe \cite{b6} and ELMo \cite{b7} in NLP community.


For evaluation, we propose MSE (Mean Squared Error) and MAE (Mean Absolute Error) as evaluation metrics for time series forecasting \cite{b19, b23}. On the other hand, for evaluating Anomaly Detection, we suggest F1 score and Recall \cite{b34}. Additionally, Accuracy is most common to evaluate time series classification \cite{b35}. By using these evaluation metrics, researchers can explore various modeling strategies and choose the best-performing models.

\section{Challenges and Limitations}

GLUE and SuperGLUE are widely used benchmarks for Natural Language Understanding (NLU) tasks. However, these benchmarks include some private test data that is not publicly available to the participants. The main reason for this is to prevent overfitting or cheating on the test sets, which could result in an unfair evaluation of the models. The private test data is derived from the same sources and domains as the public test data, but it may have different characteristics or distributions. For example, the private data for the MultiRC task may have more complex questions or answers than the public data. Private data is curated and annotated to ensure quality, but biases may still be introduced.


Time series benchmarking refers to the process of evaluating and comparing various methods used in time series analysis, such as forecasting, classification, or anomaly detection. This process is critical to advance the state-of-the-art and establish best practices in this field. However, time series benchmarking is not without its challenges. One of the main challenges is the lack of a unified and fair benchmarking framework, which can lead to inconsistencies and biases in results across different studies. Furthermore, the diversity and heterogeneity of time series data pose additional challenges, with variations in characteristics such as dimensionality, complexity, noise, seasonality, and dependencies. Selecting appropriate datasets, metrics, and baselines becomes a complex task, depending on the specific domain, task requirements, and user preferences.

Addressing these challenges requires careful design and evaluation of different time series benchmarking methods, employing various techniques such as statistical, machine learning, or deep learning approaches. To illustrate, a recent paper \cite{b30} provides a comprehensive benchmarking and heterogeneity analysis of multivariate time series forecasting. This particular problem involves predicting future values of multiple time series. The paper evaluated over 30 popular methods across more than 18 datasets, offering valuable insights for researchers to select and design effective time series forecasting methods. Additionally, common benchmark datasets may contain issues such as triviality, unrealistic anomaly density, and mislabeled ground truth. These issues are discussed in detail in this paper \cite{b31} and should be taken into account.

\section{Conclusion}

Finally, in order to benchmark time series models effectively, carefully curated datasets and evaluation metrics have been chosen, taking inspiration from natural language processing methodologies. The selected datasets cover a diverse range of domains, including electricity load forecasting, anomaly detection in areas such as news, sports, finance, entertainment, and weather, as well as time series classification across domains like image, text, and census data. However, this practice faces several challenges due to the lack of a unified benchmarking framework and the complexity and diversity of time series data. These challenges require careful consideration when selecting datasets, metrics, and baselines while acknowledging variations in data characteristics. Researchers are actively addressing these issues by designing and evaluating benchmarking approaches using statistical, machine learning, and deep learning techniques.

\section*{Acknowledgment}

I would like to express my heartfelt appreciation to Ferdinand Heinrich for his constant encouragement and insightful discussions. His expertise, feedback, and advice played a crucial role in the successful completion of this research article. 

\bibliographystyle{plain}
\bibliography{bibiliography} 

\end{document}